\documentclass[letterpaper]{article}
\usepackage[draft]{aaai2027}
\usepackage[hyphens]{url}
\usepackage{graphicx}
\usepackage{natbib}
\usepackage{caption}
\usepackage{booktabs}
\usepackage{amsmath}
\usepackage{amssymb}
\usepackage{multirow}
\usepackage{pifont}
\usepackage{placeins}
\title{When Depth Hurts: Reliability-Aware Geometry Distillation for Depth-Free RGB-D Salient Object Detection}
\author{Xuehao Wang$^{1,2}$, Jiaxin Hua$^1$, Runmei Li$^1$, Zhenyu Wu$^{4,2}$, Chenglizhao Chen$^3$, Ke Gu$^5$, Aimin Hao$^2$}
\affiliations{$^1$University of International Business and Economics, $^2$State Key Laboratory of Virtual Reality Technology and Systems,$^3$China University of Petroleum, $^4$Southwest Jiaotong University, $^5$Beijing University Of Technology}

\newcommand{\method}{GeoDistill}
\newcommand{\cmark}{\ding{51}}

\begin{document}
\maketitle

\begin{abstract}
Depth can resolve appearance ambiguity in RGB-D salient object detection (SOD), yet sensor depth is not uniformly reliable. Missing regions, blurred boundaries, and structural artifacts can propagate through multimodal fusion and make an RGB-D detector less accurate than its RGB-only counterpart. Existing quality-aware approaches regulate observed depth but remain dependent on the same potentially defective modality. We propose \method, a reliability-aware geometry distillation framework developed for RGB-D SOD benchmarks without using dataset-provided depth during training or inference. A frozen Depth Anything V2 model serves only as a training-time teacher, transferring dense relative geometry, hierarchical spatial attention, and boundary structure to a compact edge-aware geometry branch. Pooled bidirectional interaction aligns geometry with appearance, and a pixel-wise reliability estimator selectively injects geometry that is compatible with the current RGB representation. The teacher is removed after training, leaving an RGB-only inference network. Trained on 2,985 RGB-mask pairs, \method{} achieves the best or tied-best result in 26 of 36 metric-dataset comparisons against ten recent RGB-D SOD methods, including a 13.4\% relative MAE reduction on ReDWeb-S. When retrained on DUTS-TR, it also improves the strongest prior $F$-measure by 4.2\% on PASCAL-S, showing that the distilled geometry transfers beyond a particular sensor or dataset domain. Code will be released upon publication.
\end{abstract}

\section{Introduction}
Salient object detection aims to localize the most visually distinctive objects in a scene and supports image understanding, editing, retrieval, and segmentation. Modern RGB SOD models have advanced through multi-scale aggregation, boundary supervision, feedback refinement, and transformer-based context modeling \citep{borji2015salient,wang2022sodsurvey,qin2019basnet,zhao2019egnet,wei2020f3net,wang2023menet}. RGB-D SOD further uses depth to resolve appearance ambiguity: reliable geometric discontinuities can separate objects with similar colors or textures and improve boundary localization \citep{peng2014rgbd,chen2018pcf,fu2020jldcf,pang2020hdfnet}.

\begin{figure}[t]
    \centering
    \includegraphics[width=\columnwidth]{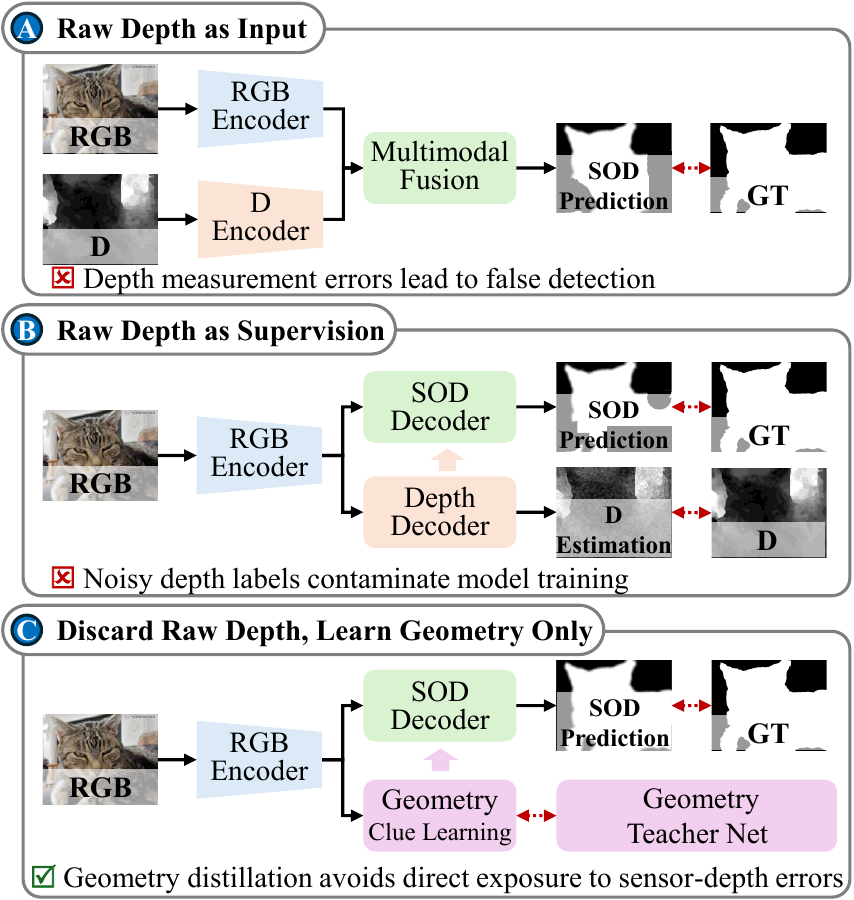}
    \caption{Three paradigms for using geometry in RGB-D SOD. Raw depth is either processed as a parallel input (A) or used to supervise a monocular-depth branch (B); measurement errors can contaminate fusion or optimization in both cases. \method{} discards dataset depth and distills task-oriented geometry from a frozen teacher (C), which is removed at inference.}
    \label{fig:motivation}
\end{figure}

The common assumption that depth is always beneficial is fragile. Public RGB-D benchmarks combine measurements from heterogeneous sensors and reconstruction pipelines, so depth quality varies markedly across scenes. Missing values, foreground-background bleeding, weak contrast, and structural artifacts can contaminate appearance features once they enter a tightly coupled fusion network. Existing methods alleviate this problem through uncertainty modeling, quality calibration, depth filtering, or selective fusion \citep{zhang2020ucnet,ji2021calibrated,fan2021rethinking}. These strategies improve how observed depth is consumed, but do not remove the model's dependence on its quality and availability.

Figure~\ref{fig:motivation} summarizes the resulting gap. Conventional RGB-D SOD either treats raw depth as a parallel input or uses it to supervise a monocular-depth branch coupled with saliency learning. The first paradigm is vulnerable to fusion-time corruption; the second transfers the same measurement errors into optimization. Removing raw depth avoids both failure modes but raises two questions: how can an RGB-only network acquire geometry informative for saliency, and how can it prevent geometrically valid yet saliency-irrelevant structures from dominating prediction?

We address these questions by separating \emph{geometry acquisition} from \emph{geometry utilization}. During training, a frozen Depth Anything V2 teacher transfers dense relative depth, multi-scale spatial attention, and boundary structure to a compact geometry branch that shares the RGB pyramid. The learned geometry exchanges context with appearance through memory-efficient bidirectional attention, after which a pixel-wise reliability estimator controls its contribution to the saliency representation. Distillation determines what geometry is learned; reliability-aware fusion determines when that geometry should be trusted.

We call this formulation \emph{depth-free RGB-D SOD}: the model follows established RGB-D benchmarks and comparison protocols but never reads their sensor depth maps. The final network requires only RGB input and does not retain the depth teacher. Our contributions are threefold:
\begin{itemize}
    \item We formulate depth-free RGB-D SOD to directly address depth-induced negative transfer, excluding dataset depth from both optimization and inference.
    \item We develop multi-level geometry distillation that transfers relative depth, hierarchical attention, and boundary structure into an edge-aware student, preserving useful geometry after the foundation teacher is removed.
    \item We introduce cross-modal enhancement and reliability-aware fusion to regulate geometry at each scale. Extensive comparisons on nine RGB-D and four RGB benchmarks, including a direct study against pseudo-depth substitution, verify the effectiveness and transferability of the framework.
\end{itemize}

\begin{figure*}[t]
    \centering
    \includegraphics[width=\textwidth]{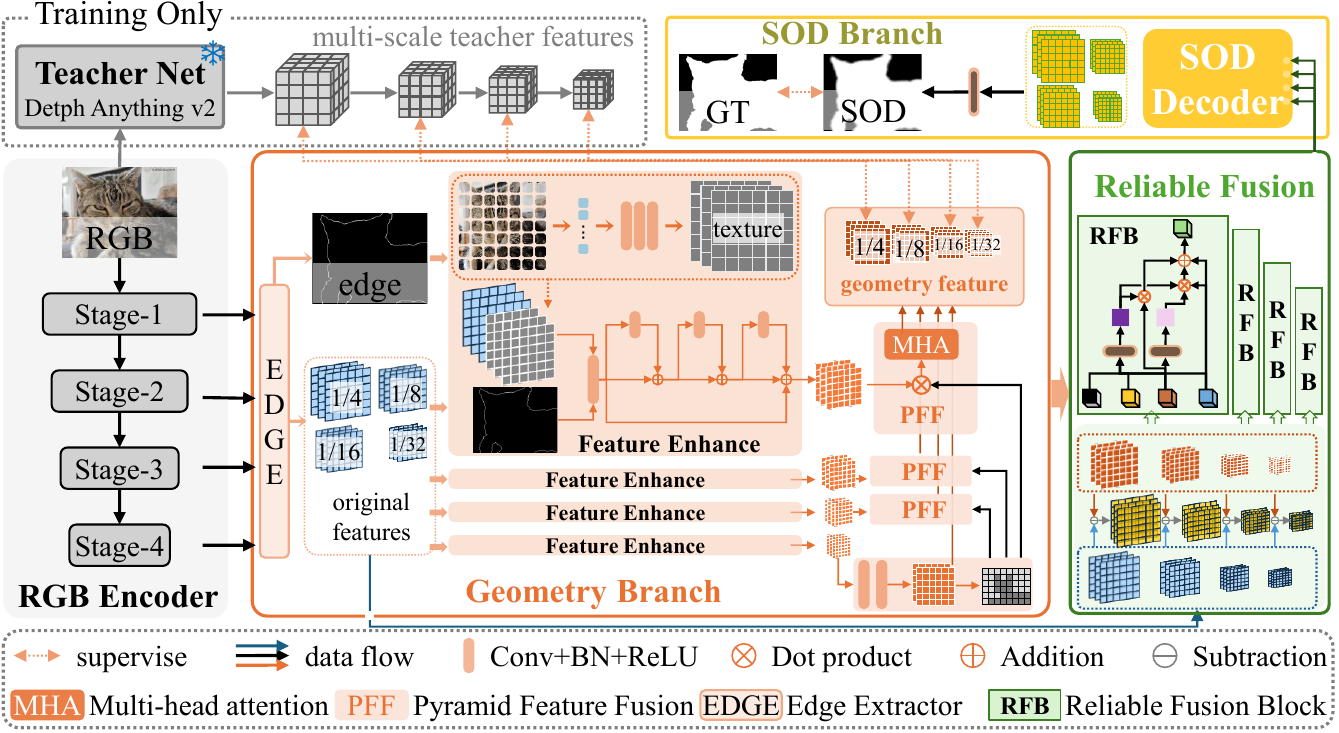}
    \caption{Architecture of \method. A shared encoder forms a four-level appearance pyramid. During training, Depth Anything V2 supervises the geometry branch at the prediction, feature, and boundary levels. Cross-modal enhancement and reliability-aware fusion produce a geometry-calibrated pyramid for the SOD decoder. Dashed teacher paths are removed after training.}
    \label{fig:framework}
\end{figure*}

\section{Related Work}
\textbf{RGB and RGB-D salient object detection.}
RGB SOD combines contextual reasoning with structure preservation through attention, partial decoding, boundary supervision, feedback, and object-level enhancement \citep{liu2018picanet,wu2019cpd,qin2019basnet,zhao2019egnet,wei2020f3net,wang2023menet}. RGB-D methods extend these designs with dual-stream or progressive cross-modal interaction, including PCFNet, JL-DCF, BBS-Net, HDFNet, and RD3D \citep{chen2018pcf,fu2020jldcf,fan2020bbsnet,pang2020hdfnet,chen2021rd3d}. In contrast, our geometry is learned from RGB rather than supplied as a sensor modality.

\textbf{Unreliable or unavailable depth.}
Prior work models depth uncertainty, calibrates modality quality, filters unreliable maps, distills depth for efficient inference, or removes depth at test time \citep{zhang2020ucnet,ji2021calibrated,fan2021rethinking,piao2020a2dele,zhang2022withoutdepth}. Recent transformer, self-supervised, and diffusion frameworks strengthen multimodal interaction \citep{zhao2022sslsod,pang2023caver,wu2023hidanet,zhang2025dimsod}. SATNet \citep{satnet} further replaces sensor depth with a monocularly estimated depth prior, but still treats the final single-channel map as an explicit input modality. In contrast, \method{} distills hierarchical teacher geometry into a compact student and removes both dataset depth and the teacher at inference.

\textbf{Foundation geometry and distillation.}
Knowledge distillation transfers prediction, feature, or attention knowledge from a high-capacity teacher to a compact student \citep{hinton2015distilling,romero2015fitnets,zagoruyko2017attention}. Self-supervised encoders and depth foundation models provide transferable scene geometry \citep{oquab2024dinov2,yang2024depthanything,yang2024depthanythingv2}. We use Depth Anything V2 as a frozen training-only teacher and transfer dense values, hierarchical attention, and boundaries rather than regressing only a pseudo-depth map. AETP and ESC-style operators from ESCNet \citep{ye2025escnet} serve as established edge-aware decoding blocks; our contribution is their integration into a geometry student and reliability-controlled SOD framework.

\section{Method}
\subsection{Overview}
As shown in Figure~\ref{fig:framework}, \method{} follows three stages: \emph{appearance encoding, geometry acquisition, and reliability-controlled saliency prediction}. A shared encoder constructs a four-level RGB pyramid and projects all levels to a common width. A geometry branch converts this pyramid into hierarchical geometry features and a dense relative-geometry map. During training, a frozen Depth Anything V2 teacher supervises the branch at the value, feature, and boundary levels; all teacher paths disappear after optimization.

The learned geometry is not sent directly to the saliency decoder. At each scale, appearance and geometry first exchange contextual information. A reliability estimator then evaluates their agreement and controls geometry injection. The fused pyramid is decoded by a separate edge-aware SOD branch. The teacher supplies transferable geometry, the student adapts it to SOD, and reliability-aware fusion suppresses structures that are geometrically plausible but irrelevant to saliency.

\subsection{Shared Pyramid and Common Projection}
Let $I\in\mathbb{R}^{3\times H\times W}$ denote an RGB image. The backbone produces four feature levels $\{X_i\}_{i=1}^{4}$ at strides $\{4,8,16,32\}$. Because ResNet-50, PVT-v2, and Swin-B expose different channel configurations, each level is transformed by an independent $1\times1$ convolution, batch normalization, and ReLU:
\begin{equation}
A_i=P_i(X_i), \qquad A_i\in\mathbb{R}^{C\times H_i\times W_i}.
\end{equation}
The common width $C$ controls the capacity of geometry learning, cross-modal interaction, and saliency decoding. We use $C=128$ by default and study $C\in\{64,128,256\}$ in the capacity ablation.

\subsection{Teacher-Guided Geometry Learning}
\textbf{Training-only teacher.}
A frozen Depth Anything V2 teacher $\mathcal{T}$ receives the same RGB image after teacher-specific resizing and normalization. We expose four DINOv2 intermediate layers and the DPT depth head:
\begin{equation}
(D^T,\{T_i\}_{i=1}^{4})=\mathcal{T}(I).
\end{equation}
$D^T$ is treated as relative geometry and normalized independently per image. Teacher parameters are never updated.

\textbf{Geometry branch.}
The student branch $\mathcal{G}$ receives the image and projected pyramid. An AETP edge extractor followed by an ESC decoder stack produces intermediate geometry predictions, four geometry features, and a geometry-edge logit:
\begin{equation}
(\{D_k\}_{k=1}^{K},\{G_i\}_{i=1}^{4},E^G)=\mathcal{G}(I,\{A_i\}).
\end{equation}
AETP combines shallow details with the deepest semantics and uses deformable convolution and self-attention to infer geometry boundaries. The ESC decoder employs image-patch references, edge-conditioned deformable sampling, multi-kernel enhancement, and coarse-to-fine feedback. These operators are inherited from ESCNet \citep{ye2025escnet} and adapted from camouflage-mask decoding to geometry learning.

\textbf{Dense relative-depth supervision.}
After min-max normalization, teacher depth $\widetilde D^T$ supervises every decoder prediction through value and gradient consistency:
\begin{equation}
\mathcal{L}_{d}=\sum_{k=1}^{K}\omega_k\left(\|\sigma(D_k)-\widetilde D_k^T\|_1+\eta\,\mathcal{L}_{\nabla}(\sigma(D_k),\widetilde D_k^T)\right),
\end{equation}
where $\omega_k$ emphasizes later outputs and $\mathcal{L}_{\nabla}$ measures horizontal and vertical gradient discrepancies.

\textbf{Hierarchical feature alignment.}
Because teacher and student features differ architecturally, we align their normalized channel-energy maps rather than raw tensors. For feature $F$, define
\begin{equation}
\mathcal{A}(F)=\mathcal{N}\left(\frac{1}{C_F}\sum_{c=1}^{C_F}F_c^2\right),
\end{equation}
where $\mathcal{N}$ denotes spatial min-max normalization. The alignment objective is
\begin{equation}
\mathcal{L}_{a}=\frac{1}{4}\sum_{i=1}^{4}\|\mathcal{A}(G_i)-\mathcal{A}(T_i)\|_1.
\end{equation}
This transfers the teacher's spatial focus while allowing task-specific student channels.

\textbf{Geometry-boundary supervision.}
The gradient-derived boundary of $\widetilde D^T$ supervises $E^G$:
\begin{equation}
\mathcal{L}_{ge}=\operatorname{BCE}(E^G,\operatorname{Edge}(\widetilde D^T)).
\end{equation}
Thus, $\mathcal{L}_{d}$, $\mathcal{L}_{a}$, and $\mathcal{L}_{ge}$ transfer value-, feature-, and boundary-level geometry knowledge.

\subsection{Cross-Modal Enhancement}
Appearance and geometry originate from the same image but encode different inductive biases. At each level, native-resolution queries attend to adaptively pooled keys and values, reducing spatial attention from quadratic complexity to $O(H_iW_iP^2)$ for pooled size $P\times P$:
\begin{align}
\bar A_i &= A_i + \alpha_i\operatorname{Attn}(Q_A(A_i),K_G(\Pi(G_i)),V_G(\Pi(G_i))),\\
\bar G_i &= G_i + \beta_i\operatorname{Attn}(Q_G(G_i),K_A(\Pi(A_i)),V_A(\Pi(A_i))),
\end{align}
where $\Pi$ is adaptive average pooling. Residual gates $\alpha_i$ and $\beta_i$ are initialized to zero so training starts from the independent branches. Residual coordinate attention further captures horizontal and vertical dependencies \citep{hou2021coordinate}.

\subsection{Reliability-Aware Geometry Fusion}
Monocular geometry may describe walls, ground planes, or background discontinuities that are valid in 3D but irrelevant to saliency. We therefore estimate a pixel-wise reliability map instead of assigning geometry a fixed contribution. After modality-specific projection, the estimator receives appearance, geometry, their absolute discrepancy, and the resized geometry prediction:
\begin{equation}
r_i=\sigma\left(\phi_i([\bar A_i,\bar G_i,|\bar A_i-\bar G_i|,\mathcal{U}_i(\sigma(D))])\right).
\end{equation}
A channel-wise geometry attention map is $q_i=\sigma(\psi_i(\bar G_i))$, and fusion is
\begin{equation}
F_i=\rho_i\left(\bar A_i\odot(1+r_i\odot q_i)+r_i\odot\bar G_i\right).
\end{equation}
This RGB-dominant formulation approaches the appearance baseline when $r_i$ is small and activates multiplicative modulation and residual geometry injection when the branches agree. A negative bias in the final reliability layer prevents unstable geometry from dominating early optimization.

\subsection{Edge-Aware Saliency Decoding}
A separate AETP and ESC decoder stack transforms $\{F_i\}$ into saliency logits $\{S_j\}_{j=1}^{J}$ and an SOD edge logit $E^S$. Geometry and saliency decoding share the same progressive structure but not parameters, allowing one branch to preserve teacher geometry and the other to optimize foreground selection. The saliency branch uses structure loss with deep supervision:
\begin{equation}
\mathcal{L}_{s}=\sum_{j=1}^{J}\nu_j\mathcal{L}_{\mathrm{str}}(S_j,Y), \qquad
\mathcal{L}_{e}=\operatorname{BCE}(E^S,\operatorname{Edge}(Y)).
\end{equation}
The complete objective is
\begin{equation}
\mathcal{L}=\mathcal{L}_{s}+\lambda_e\mathcal{L}_{e}+\lambda_{ge}\mathcal{L}_{ge}+\lambda_d\mathcal{L}_{d}+\lambda_a\mathcal{L}_{a}.
\end{equation}
We set $\lambda_e=0.4$, $\lambda_{ge}=0.1$, $\lambda_d=0.2$, and $\lambda_a=0.05$. At inference, $\mathcal{T}$ and all supervision paths are removed.

\section{Experiments}
\subsection{Experimental Protocol}
\textbf{Datasets.}
For RGB-D SOD, we train on 2,985 RGB-mask pairs: 1,485 from NJU2K \citep{ju2015depthaware}, 700 from NLPR \citep{peng2014rgbd}, and 800 from DUT-RGBD \citep{piao2019dimin}. Dataset depth is ignored. Evaluation uses NJU2K (500 test images), NLPR (300), DUT-RGBD (400), ReDWeb-S (1,000) \citep{liu2021smac}, SIP (929) \citep{fan2021rethinking}, SSD (80) \citep{zhou2021rgbdsurvey}, STERE (1,000) \citep{niu2012stereopsis}, COME-E (4,600), and COME-H (3,000) \citep{zhang2021cmi}. For RGB-only generalization, we retrain on the 10,553-image DUTS training split and evaluate on DUTS-TE (5,019) \citep{wang2017duts}, ECSSD (1,000) \citep{shi2016hierarchical}, HKU-IS (4,447) \citep{li2016multiscale}, and PASCAL-S (850) \citep{li2014secrets}.

\textbf{Metrics.}
We report structure measure $S_m$ \citep{fan2017structure}, maximum F-measure $F_\beta^{\max}$ with $\beta^2=0.3$, maximum enhanced-alignment measure $E_\xi^{\max}$ \citep{fan2018enhanced}, and mean absolute error $\mathcal{M}$. Higher values are better for the first three metrics, whereas lower $\mathcal{M}$ is better.

\textbf{Implementation details.}
Unless stated otherwise, we use PVT-v2-B5 as the encoder, Depth Anything V2-Small as the frozen teacher, projector width $C=128$, and one ESC block in each branch. Student and teacher inputs are resized to $416\times416$ and $364\times364$, respectively. We apply random cropping and horizontal flipping. Training lasts 80 epochs with batch size 4 and AdamW. The learning rate is $7.5\times10^{-5}$ for newly initialized modules and $7.5\times10^{-6}$ for the encoder; weight decay is $1.5\times10^{-4}$ and gradients are clipped to 1.0. Training is conducted on a single RTX 4090 GPU with 24 GB memory.

\begin{table*}[t]
\centering
\caption{Quantitative comparison on nine RGB-D SOD benchmarks. Numbers in parentheses are test-set sizes. Best and tied-best results are bold.}
\label{tab:rgbd_sota}
\resizebox{\textwidth}{!}{%
\begin{tabular}{l|cccc|cccc|cccc|cccc|cccc}
\toprule
Method & \multicolumn{4}{c|}{NJU2K (500)} & \multicolumn{4}{c|}{NLPR (300)} & \multicolumn{4}{c|}{DUT-RGBD (400)} & \multicolumn{4}{c|}{ReDWeb-S (1,000)} & \multicolumn{4}{c}{SIP (929)} \\
\cmidrule(lr){2-5}\cmidrule(lr){6-9}\cmidrule(lr){10-13}\cmidrule(lr){14-17}\cmidrule(l){18-21}
& $S_m$ & $F_\beta^{\max}$ & $E_\xi^{\max}$ & $\mathcal{M}$ & $S_m$ & $F_\beta^{\max}$ & $E_\xi^{\max}$ & $\mathcal{M}$ & $S_m$ & $F_\beta^{\max}$ & $E_\xi^{\max}$ & $\mathcal{M}$ & $S_m$ & $F_\beta^{\max}$ & $E_\xi^{\max}$ & $\mathcal{M}$ & $S_m$ & $F_\beta^{\max}$ & $E_\xi^{\max}$ & $\mathcal{M}$ \\
\midrule
C2DFNet & .861 & .854 & .912 & .054 & .909 & .895 & .953 & .025 & .896 & .904 & .942 & .037 & .613 & .580 & .708 & .175 & .793 & .784 & .855 & .088 \\
RD3D & .893 & .883 & .927 & .047 & .903 & .880 & .937 & .033 & .863 & .841 & .894 & .060 & .671 & .621 & .730 & .163 & .835 & .826 & .884 & .074 \\
PICRNet & .386 & .274 & .526 & .423 & .389 & .157 & .609 & .383 & .362 & .247 & .515 & .432 & .356 & .311 & .486 & .447 & .312 & .246 & .571 & .464 \\
HRTransNet & .917 & .920 & .952 & .032 & .931 & .925 & .966 & .019 & .918 & .925 & .951 & .033 & .724 & .710 & .800 & .127 & .860 & .876 & .916 & .056 \\
CAVER & .926 & .928 & .959 & .030 & .934 & .929 & .970 & .021 & .938 & .944 & .966 & .026 & .736 & .737 & .808 & .121 & .904 & .915 & .945 & .038 \\
CPNet & \textbf{.935} & \textbf{.941} & \textbf{.964} & \textbf{.025} & \textbf{.940} & \textbf{.936} & \textbf{.973} & \textbf{.016} & .951 & \textbf{.959} & .975 & \textbf{.019} & .752 & .755 & .822 & .112 & .907 & .927 & .946 & .035 \\
LAFB & .907 & .912 & .946 & .036 & .930 & .921 & .965 & .020 & .927 & .934 & .956 & .028 & .722 & .721 & .792 & .129 & .897 & .913 & .942 & .041 \\
CATNet & .932 & .937 & .961 & .026 & \textbf{.940} & .934 & .972 & .018 & \textbf{.953} & .958 & \textbf{.976} & \textbf{.019} & .748 & .750 & .816 & .115 & .911 & \textbf{.928} & \textbf{.952} & .034 \\
SATNet & .923 & .925 & .954 & .030 & .929 & .920 & .964 & .021 & .942 & .947 & .966 & .022 & .705 & .701 & .782 & .131 & .898 & .904 & .931 & .042 \\
DPPNet & .929 & .932 & .962 & .028 & .937 & .927 & .968 & .020 & .939 & .946 & .965 & .025 & .749 & .746 & .817 & .115 & .896 & .911 & .938 & .042 \\
\textbf{\method} & \textbf{.935} & .940 & \textbf{.964} & \textbf{.025} & .935 & .928 & .967 & .019 & .948 & .957 & .973 & .020 & \textbf{.781} & \textbf{.789} & \textbf{.842} & \textbf{.097} & \textbf{.912} & \textbf{.928} & .951 & \textbf{.033} \\
\bottomrule
\end{tabular}}

\smallskip
\resizebox{0.86\textwidth}{!}{%
\begin{tabular}{l|cccc|cccc|cccc|cccc}
\toprule
Method & \multicolumn{4}{c|}{SSD (80)} & \multicolumn{4}{c|}{STERE (1,000)} & \multicolumn{4}{c|}{COME-E (4,600)} & \multicolumn{4}{c}{COME-H (3,000)} \\
\cmidrule(lr){2-5}\cmidrule(lr){6-9}\cmidrule(lr){10-13}\cmidrule(l){14-17}
& $S_m$ & $F_\beta^{\max}$ & $E_\xi^{\max}$ & $\mathcal{M}$ & $S_m$ & $F_\beta^{\max}$ & $E_\xi^{\max}$ & $\mathcal{M}$ & $S_m$ & $F_\beta^{\max}$ & $E_\xi^{\max}$ & $\mathcal{M}$ & $S_m$ & $F_\beta^{\max}$ & $E_\xi^{\max}$ & $\mathcal{M}$ \\
\midrule
C2DFNet & .807 & .762 & .872 & .067 & .868 & .863 & .919 & .048 & .779 & .776 & .848 & .094 & .724 & .726 & .800 & .130 \\
RD3D & .852 & .814 & .901 & .058 & .889 & .868 & .920 & .048 & .836 & .818 & .877 & .073 & .782 & .764 & .824 & .109 \\
PICRNet & .344 & .253 & .444 & .467 & .386 & .253 & .539 & .421 & .362 & .295 & .489 & .427 & .361 & .329 & .483 & .434 \\
HRTransNet & .848 & .820 & .909 & .053 & .915 & .912 & .954 & .032 & .858 & .856 & .909 & .056 & .816 & .818 & .869 & .084 \\
CAVER & .890 & .884 & .936 & .039 & .917 & .916 & .955 & .033 & .870 & .874 & .918 & .052 & .822 & .831 & .872 & .082 \\
CPNet & .893 & .893 & .935 & .035 & .920 & .923 & \textbf{.960} & .029 & .884 & .889 & .928 & .045 & .843 & .854 & .889 & .071 \\
LAFB & .857 & .841 & .922 & .045 & .908 & .906 & .945 & .037 & .864 & .864 & .908 & .056 & .814 & .817 & .860 & .088 \\
CATNet & .892 & .879 & .927 & .036 & .921 & .922 & .958 & .030 & .892 & .897 & .932 & .043 & .847 & .855 & .890 & .071 \\
SATNet & .871 & .852 & .917 & .044 & .919 & .913 & .951 & .032 & .864 & .855 & .901 & .056 & .814 & .809 & .857 & .087 \\
DPPNet & .891 & .885 & .938 & .037 & .922 & .919 & .957 & .032 & .878 & .876 & .917 & .052 & .839 & .840 & .879 & .078 \\
\textbf{\method} & \textbf{.898} & \textbf{.894} & \textbf{.948} & \textbf{.031} & \textbf{.927} & \textbf{.925} & \textbf{.960} & \textbf{.028} & \textbf{.896} & \textbf{.902} & \textbf{.935} & \textbf{.041} & \textbf{.856} & \textbf{.869} & \textbf{.899} & \textbf{.065} \\
\bottomrule
\end{tabular}}
\end{table*}

\FloatBarrier

\subsection{Quantitative Comparison on RGB-D SOD}
We compare \method{} with ten recent RGB-D SOD methods: C2DFNet \citep{miao2022c2dfnet}, RD3D \citep{chen20223}, PICRNet \citep{cong2023point}, HRTransNet \citep{hrtransnet}, CAVER \citep{pang2023caver}, CPNet \citep{hu2024progressive}, LAFB \citep{wang2024learning}, CATNet \citep{sun2024catnet}, SATNet \citep{satnet}, and DPPNet \citep{dppnet}. Table~\ref{tab:rgbd_sota} reports all four metrics on nine benchmarks; numbers in parentheses denote evaluated test images.

Across the 36 metric-dataset comparisons, \method{} is best or tied-best in 26 cases (72.2\%), including 21 outright best results. Its advantage is most pronounced on benchmarks that differ substantially from the training distribution. On ReDWeb-S, the strongest prior $S_m$, $F_\beta^{\max}$, and $E_\xi^{\max}$ are improved by 3.9\%, 4.5\%, and 2.4\%, respectively, while MAE decreases by 13.4\%. MAE is also reduced by 11.4\% on SSD, 8.5\% on COME-H, 4.7\% on COME-E, and 3.5\% on STERE. On NJU2K, \method{} ties the best $S_m$, $E_\xi^{\max}$, and MAE, and its $F_\beta^{\max}$ is within 0.1\% of the top result. Performance on NLPR and DUT-RGBD remains competitive but is not uniformly best. Overall, the results support a precise conclusion: distilled geometry is particularly robust to cross-dataset variation and heterogeneous depth quality.

\begin{figure*}[t]
    \centering
    \includegraphics[width=\textwidth]{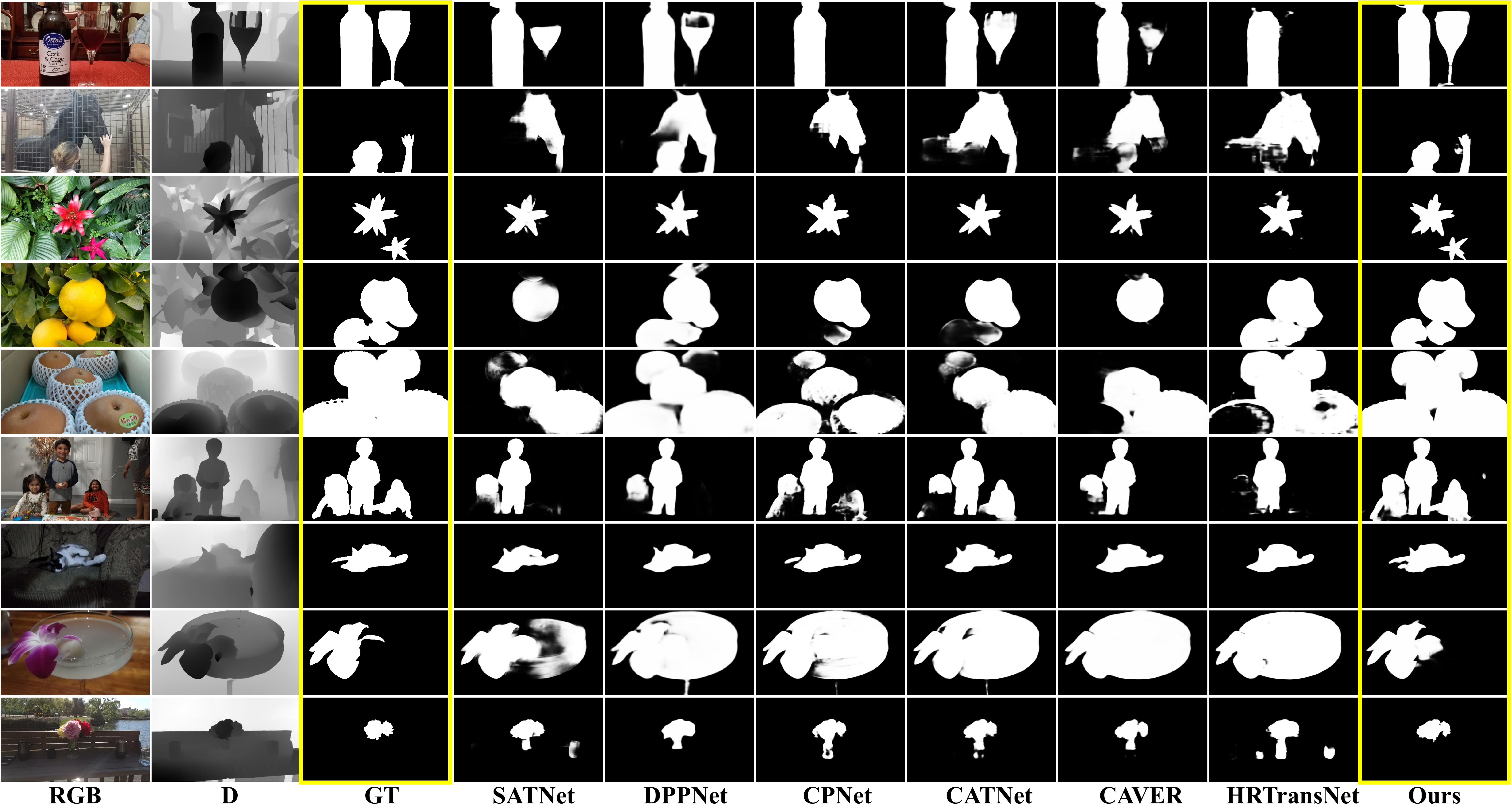}
    \caption{Qualitative comparison. Rows cover three representative depth conditions: informative but RGB-ambiguous geometry, depth maps contaminated by structured background, and incomplete or blurred depth boundaries. \method{} preserves complete salient regions and suppresses depth-induced false positives across all three conditions.}
    \label{fig:qualitative}
\end{figure*}

Figure~\ref{fig:qualitative} groups challenging cases by depth condition. When RGB appearance is ambiguous but geometry is informative, \method{} recovers complete foreground regions. When raw depth contains structured background responses, it suppresses the false positives and missed objects produced by competing models. When depth boundaries are incomplete or blurred, it preserves object contours more consistently. These examples illustrate the benefit of learning geometry from RGB and regulating its contribution instead of directly consuming sensor depth.

\subsection{Comparison with Pseudo-Depth Substitution}
\begin{figure}[t]
    \centering
    \includegraphics[width=\columnwidth]{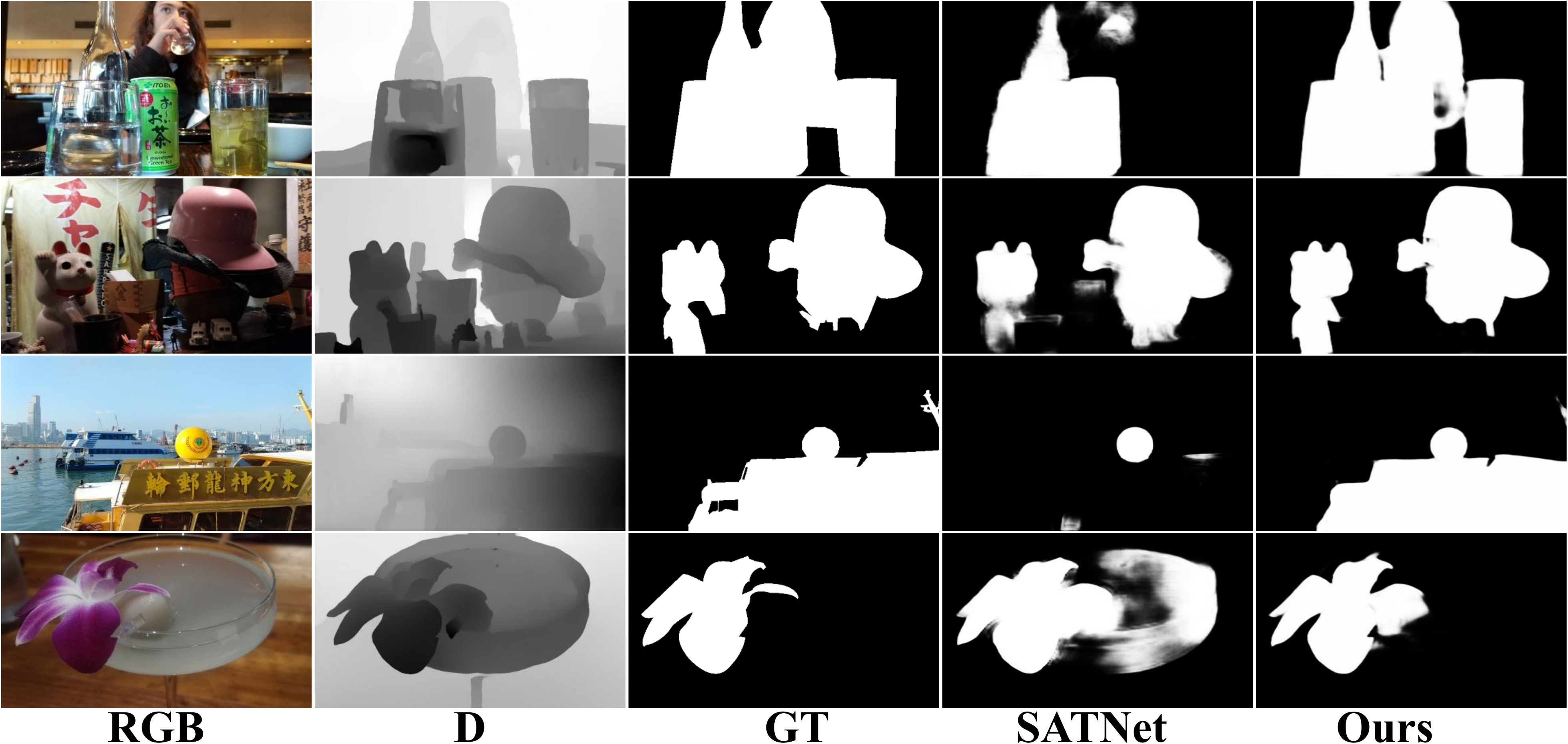}
    \caption{Comparison with SATNet. SATNet replaces sensor depth with a monocular pseudo-depth input, whereas \method{} distills hierarchical teacher geometry and produces more complete masks with fewer background responses.}
    \label{fig:qualitative-satnet}
\end{figure}

SATNet \citep{satnet} also avoids direct use of sensor depth, but follows a different strategy: it replaces raw depth with a monocularly estimated single-channel prior and processes RGB and pseudo-depth through symmetric input streams. By contrast, \method{} uses Depth Anything V2 only during training and distills its dense prediction, multi-scale representations, and boundaries into an internal geometry branch. Table~\ref{tab:rgbd_sota} shows that \method{} exceeds SATNet in all 36 metric-dataset comparisons; relative MAE reductions reach 29.5\% on SSD, 26.8\% on COME-E, 26.0\% on ReDWeb-S, and 25.3\% on COME-H. Figure~\ref{fig:qualitative-satnet} provides representative examples, where hierarchical geometry distillation yields more complete objects and fewer background responses than final-map substitution. These results suggest that preserving multi-scale teacher structure is more effective than compressing geometry into a single pseudo-depth input.

\subsection{Generalization to RGB SOD}
We retrain the same architecture on DUTS-TR while retaining Depth Anything V2 only as a geometry teacher. As shown in Table~\ref{tab:rgb_sota}, the framework attains the best result in nine of twelve metric-dataset comparisons. On PASCAL-S, it improves the strongest prior $S_m$, $F_\beta^{\max}$, and $E_\xi^{\max}$ by 1.3\%, 4.2\%, and 3.3\%, respectively; on HKU-IS, all three metrics improve by approximately 0.5\%. Gains on DUTS-TE are smaller but consistent (0.2--0.4\%), while all ECSSD results remain within 0.2\% of the best. The geometry student therefore learns a transferable structural prior rather than a sensor- or dataset-specific shortcut.

\begin{table*}[t]
\centering
\caption{Comparison with RGB SOD methods. Numbers in parentheses are test-set sizes. Best results are bold.}
\label{tab:rgb_sota}
\resizebox{0.88\textwidth}{!}{%
\begin{tabular}{l|ccc|ccc|ccc|ccc}
\toprule
Method & \multicolumn{3}{c|}{DUTS-TE (5,019)} & \multicolumn{3}{c|}{ECSSD (1,000)} & \multicolumn{3}{c|}{HKU-IS (4,447)} & \multicolumn{3}{c}{PASCAL-S (850)} \\
\cmidrule(lr){2-4}\cmidrule(lr){5-7}\cmidrule(lr){8-10}\cmidrule(l){11-13}
& $S_m$ & $F_\beta^{\max}$ & $E_\xi^{\max}$ & $S_m$ & $F_\beta^{\max}$ & $E_\xi^{\max}$ & $S_m$ & $F_\beta^{\max}$ & $E_\xi^{\max}$ & $S_m$ & $F_\beta^{\max}$ & $E_\xi^{\max}$ \\
\midrule
VST \citep{liu2021vst} & .896 & .877 & .939 & .932 & .944 & .964 & .928 & .937 & .968 & .873 & .850 & .900 \\
ICON \citep{zhuge2023icon} & .890 & .876 & .931 & .928 & .943 & .960 & .920 & .931 & .960 & .862 & .844 & .888 \\
VST-T++ \citep{liu2023vstpp} & .901 & .887 & .943 & .937 & .949 & .968 & .930 & .939 & .968 & .878 & .855 & .901 \\
MENet \citep{wang2023menet} & .905 & .895 & .943 & .927 & .938 & .956 & .927 & .939 & .965 & .871 & .848 & .892 \\
VSCode-T \citep{luo2024vscode} & .917 & .910 & .954 & \textbf{.945} & \textbf{.957} & \textbf{.971} & .935 & .946 & .970 & .878 & .852 & .900 \\
VSCode-v2-T \citep{luo2025vscodev2} & .922 & .917 & .957 & .940 & .950 & .965 & .929 & .936 & .962 & .875 & .847 & .891 \\
\textbf{\method} & \textbf{.926} & \textbf{.920} & \textbf{.959} & .943 & .956 & .970 & \textbf{.940} & \textbf{.951} & \textbf{.975} & \textbf{.889} & \textbf{.891} & \textbf{.931} \\
\bottomrule
\end{tabular}}
\end{table*}

\subsection{Ablation Study}
All ablations use the same split, validation criterion, and evaluation protocol. We examine three questions: how much shared feature capacity is required, whether distilled geometry can replace raw depth, and whether teacher-guided geometry learning adds value beyond the branch architecture alone.

\textbf{Projector capacity.}
Table~\ref{tab:capacity} varies $C\in\{64,128,256\}$. Increasing $C$ from 64 to 128 reduces MAE by 36.5\% on SIP and 24.4\% on COME-H, with relative gains of up to 0.8\% in the region metrics. Increasing $C$ further to 256 does not improve accuracy: the 128-channel model is slightly better on every reported metric while using 43.9\% fewer parameters and 68.5\% fewer FLOPs. Thus, $C=64$ under-represents the geometry and saliency pyramids, whereas $C=256$ adds substantial redundancy. We adopt $C=128$ as the best accuracy-efficiency trade-off.

\begin{table}[t]
\centering
\caption{Projector-capacity ablation. Best results are bold.}
\label{tab:capacity}
\resizebox{\columnwidth}{!}{%
\begin{tabular}{c cc cccc cccc}
\toprule
$C$ & Params & FLOPs & \multicolumn{4}{c}{SIP} & \multicolumn{4}{c}{COME-H} \\
\cmidrule(lr){4-7}\cmidrule(l){8-11}
& (M) & (G) & $S_m$ & $F_\beta^{\max}$ & $E_\xi^{\max}$ & $\mathcal{M}$ & $S_m$ & $F_\beta^{\max}$ & $E_\xi^{\max}$ & $\mathcal{M}$ \\
\midrule
64  & 89.68 & 139.46 & .906 & .921 & .944 & .052 & .855 & .865 & .898 & .086 \\
128 & 111.76 & 311.34 & \textbf{.912} & \textbf{.928} & \textbf{.951} & \textbf{.033} & \textbf{.856} & \textbf{.869} & \textbf{.899} & \textbf{.065} \\
256 & 199.03 & 989.56 & .911 & .926 & .949 & .034 & .854 & .866 & .897 & .066 \\
\bottomrule
\end{tabular}}
\end{table}

\textbf{Depth-use strategy.}
Table~\ref{tab:strategy} compares three conceptually distinct settings. \emph{RGB-only} removes the geometry branch and retains only saliency and boundary supervision. \emph{Raw depth} replaces teacher-guided geometry learning with a parallel depth encoder and concatenates sensor-depth features with RGB features. \emph{Distilled geometry} is the complete depth-free model. Relative to RGB-only, distilled geometry reduces MAE by 28.3\% on SIP and 18.8\% on COME-H, while improving the region metrics by 2.1--3.3\%. It also outperforms raw-depth fusion, reducing MAE by 8.3\% and 3.0\%, respectively, with gains of up to 0.6\% in the remaining metrics. This strategic comparison is not parameter matched; it directly verifies that training-time geometry transfer can replace test-time sensor depth.

\begin{table}[t]
\centering
\caption{Depth-use strategies. Best results are bold.}
\label{tab:strategy}
\resizebox{\columnwidth}{!}{%
\begin{tabular}{l cccc cccc}
\toprule
Strategy & \multicolumn{4}{c}{SIP} & \multicolumn{4}{c}{COME-H} \\
\cmidrule(lr){2-5}\cmidrule(l){6-9}
& $S_m$ & $F_\beta^{\max}$ & $E_\xi^{\max}$ & $\mathcal{M}$ & $S_m$ & $F_\beta^{\max}$ & $E_\xi^{\max}$ & $\mathcal{M}$ \\
\midrule
RGB-only & .883 & .902 & .931 & .046 & .831 & .841 & .875 & .080 \\
Raw depth & .908 & .923 & .945 & .036 & .853 & .866 & .898 & .067 \\
Distilled geometry & \textbf{.912} & \textbf{.928} & \textbf{.951} & \textbf{.033} & \textbf{.856} & \textbf{.869} & \textbf{.899} & \textbf{.065} \\
\bottomrule
\end{tabular}}
\end{table}

\begin{table}[t]
\centering
\caption{Geometry architecture and teacher-supervision ablation. Best results are bold.}
\label{tab:geometry}
\resizebox{\columnwidth}{!}{%
\begin{tabular}{lcc cccc cccc}
\toprule
Setting & Geo. & DA-V2 & \multicolumn{4}{c}{SIP} & \multicolumn{4}{c}{COME-H} \\
\cmidrule(lr){4-7}\cmidrule(l){8-11}
& & & $S_m$ & $F_\beta^{\max}$ & $E_\xi^{\max}$ & $\mathcal{M}$ & $S_m$ & $F_\beta^{\max}$ & $E_\xi^{\max}$ & $\mathcal{M}$ \\
\midrule
RGB-only & & & .883 & .902 & .931 & .046 & .831 & .841 & .875 & .080 \\
Geometry architecture & \cmark & & .908 & .923 & .945 & .037 & .854 & .865 & .896 & .068 \\
Full model & \cmark & \cmark & \textbf{.912} & \textbf{.928} & \textbf{.951} & \textbf{.033} & \textbf{.856} & \textbf{.869} & \textbf{.899} & \textbf{.065} \\
\bottomrule
\end{tabular}}
\end{table}

\textbf{Geometry architecture and teacher supervision.}
Table~\ref{tab:geometry} separates architectural capacity from teacher guidance. \emph{RGB-only} contains the shared encoder and SOD decoder. \emph{Geometry architecture} adds the geometry branch but optimizes it only through the downstream saliency objective. \emph{Full model} further introduces Depth Anything V2 supervision at the value, feature, and boundary levels. Adding the geometry branch without teacher guidance to RGB-only reduces MAE by 19.6\% on SIP and 15.0\% on COME-H, with gains of up to 2.9\% in the region metrics. Teacher-guided geometry learning then reduces MAE by a further 10.8\% and 4.4\%, respectively. Overall, the full model lowers MAE by 28.3\% on SIP and 18.8\% on COME-H relative to RGB-only, showing that branch capacity and multi-level geometry supervision are complementary.

\subsection{Discussion}
The experiments provide complementary evidence. The nine-benchmark comparison shows that distilled geometry is particularly robust under cross-dataset variation. The direct comparison with SATNet indicates that hierarchical teacher transfer is more effective than substituting a final pseudo-depth map, while the RGB SOD results show that the benefit is not tied to an RGB-D sensor domain. The strategy, capacity, and component studies further verify that training-time geometry can replace test-time raw depth and identify the contributions of feature capacity and multi-level teacher supervision.

The geometry map represents task-oriented relative structure, not calibrated metric depth, and should not be interpreted as a replacement for a physical sensor in measurement tasks. Training also requires a frozen teacher, although it contributes no parameters or computation at inference. Caching teacher outputs or using a smaller geometry foundation model could reduce training cost.

\section{Conclusion}
We presented \method, a reliability-aware geometry distillation framework for depth-free RGB-D SOD. Instead of fusing potentially unreliable sensor depth or substituting a final pseudo-depth map, the model transfers relative depth, hierarchical attention, and boundary structure from a frozen Depth Anything V2 teacher into a compact geometry branch. Cross-modal enhancement and pixel-wise reliability estimation then determine when geometry should influence appearance. The teacher is removed after training, so inference requires only RGB. Results on nine RGB-D and four RGB benchmarks show that selectively distilled geometry is a more stable and transferable auxiliary signal than unconditional raw-depth or pseudo-depth input.

\bibliography{references}
\end{document}